\documentclass[letterpaper]{article} % DO NOT CHANGE THIS
\usepackage{aaai23}  % DO NOT 
\usepackage{times}  % DO NOT CHANGE THIS
\usepackage{helvet}  % DO NOT CHANGE THIS
\usepackage{courier}  % DO NOT CHANGE THIS
\usepackage[hyphens]{url}  % DO NOT CHANGE THIS
\usepackage{graphicx} % DO NOT CHANGE THIS

\usepackage{algorithm}
\usepackage{algorithmic}

\usepackage{multirow}

\newcommand{\ifmixup}{ifMixup}

\usepackage[colorinlistoftodos]{todonotes}

\usepackage{microtype}
\usepackage{graphicx}
\usepackage{subfigure}
\usepackage{booktabs} % for professional tables
\usepackage{amsmath}
\usepackage{amssymb}
\usepackage{mathtools}
\usepackage{amsthm}
\usepackage[capitalize,noabbrev]{cleveref}

\theoremstyle{plain}
\newtheorem{theorem}{Theorem}[section]

\newtheorem{lem}[theorem]{Lemma}

\theoremstyle{definition}

\theoremstyle{remark}

\usepackage{natbib}  % DO NOT CHANGE THIS AND DO NOT ADD ANY OPTIONS TO IT
\usepackage{caption} % DO NOT CHANGE THIS AND DO NOT ADD ANY OPTIONS TO IT
\usepackage{algorithm}
\usepackage{algorithmic}

\usepackage{newfloat}
\usepackage{listings}
\DeclareCaptionStyle{ruled}{labelfont=normalfont,labelsep=colon,strut=off} % DO NOT CHANGE THIS
\floatstyle{ruled}
\newfloat{listing}{tb}{lst}{}
\floatname{listing}{Listing}
\title{ifMixup: Interpolating Graph Pair to Regularize Graph Classification}

\author {
    Hongyu Guo\textsuperscript{\rm 1,2}, 
    Yongyi Mao \textsuperscript{\rm 2}
}
\affiliations {
    \textsuperscript{\rm 1} National Research Council Canada\\
    \textsuperscript{\rm 2} 
    University of Ottawa\\
    hongyu.guo@uottawa.ca, ymao@uottawa.ca
}

\usepackage{bibentry}
\begin{document}

\maketitle

 \begin{abstract}
 We present a simple and yet effective interpolation-based regularization technique, aiming to improve the generalization of  Graph Neural Networks (GNNs) on supervised graph classification. We leverage  Mixup~\cite{MixUp17}, an effective  regularizer for vision, where random sample pairs and their labels are interpolated to create synthetic images for training. Unlike images with grid-like coordinates, graphs have arbitrary structure and topology, which can be very sensitive to any modification that alters the graph's  semantic meanings. This posts two unanswered questions for Mixup-like regularization schemes:  
 Can we directly mix up a pair of graph inputs? 
  If so, how well does such mixing strategy  regularize the learning of GNNs?  To answer these two questions, we propose \textit{\ifmixup}, which first adds  dummy nodes  to make two graphs have the same input size and then simultaneously   performs linear interpolation between  the aligned node feature vectors and the aligned edge representations of the two graphs. 
  We empirically show that such simple mixing schema can effectively regularize the  classification learning, resulting in superior predictive accuracy to popular graph augmentation and GNN methods. 
 \end{abstract}
 
\section{Introduction}
Graph Neural Networks (GNNs)~\cite{kipf2017semi}  have recently shown  promising performance in many challenging  applications, including predicting molecule property~\cite{moleculenet}, forecasting protein activation~\cite{aptrank}, and estimating circuit functionality~\cite{circuit-gnn}. Nevertheless, like other successfully deployed deep  networks, GNNs  also suffer from the data-hungry issue due to their over-parameterized learning paradigm. Consequently, 
regularization techniques have been actively proposed, aiming to empower the learning of GNNs while avoiding over-smoothing~\cite{LiHW18}, over-squashing~\cite{alon2021on} and  over-fitting~\cite{10.1145/3446776}. 
Amongst these generalization strategies, data augmentation schemes have been proven to  be very effective~\cite{rong2020dropedge,3412086zhou}. Existing graph data augmentation methods, nevertheless,    
mostly involve graph manipulations  on a {\em single graph}~\cite{rong2020dropedge,3412086zhou,You2020GraphCL}. 

In this paper, we  look into a very successful {\em pairwise data} augmentation technique for image recognition~\cite{MixUp17,ManifoldMixUp,GuoMZ19,kimICML20}, % and natural text classification~\cite{Guo_2020,jindal-etal-2020-augmenting},
called Mixup.  
Mixup  was originally introduced by~\cite{MixUp17}   as an interpolation-based regularizer for  image classification. It regularizes the learning of deep classification models by training with  synthetic samples, which are created by  linearly  interpolating  a pair of randomly selected training samples, naturally well-aligned,  as well as their training targets.

Motivated by the effectiveness and simplicity of Mixup in regularizing image
classification models, we are motivated to design a similar, straight-forward, “Mixup” scheme for graph data. 
Nevertheless, 
unlike images, which are supported on  grid coordinates,
graph data have arbitrary structure and topology, such as 
having different numbers of nodes and these nodes from different graphs are typically not readily aligned. Furthermore, such irregular  graph topologies
typically  play a critical role in the graph semantics, and consequently even simply modifying one node or edge  can  dramatically change the semantic meaning of a graph. 
Hence, 
the following two questions naturally arise: 

\vspace{1mm}
 \small{
 \noindent \textit{\textbf{Can we directly mix up a pair of graph inputs?
  If so, how well does such mixing strategy  regularize the learning of GNNs?}}
}
\vspace{1mm}

\begin{figure*}%[h]
	\centering
	{\includegraphics[width=0.9\textwidth]{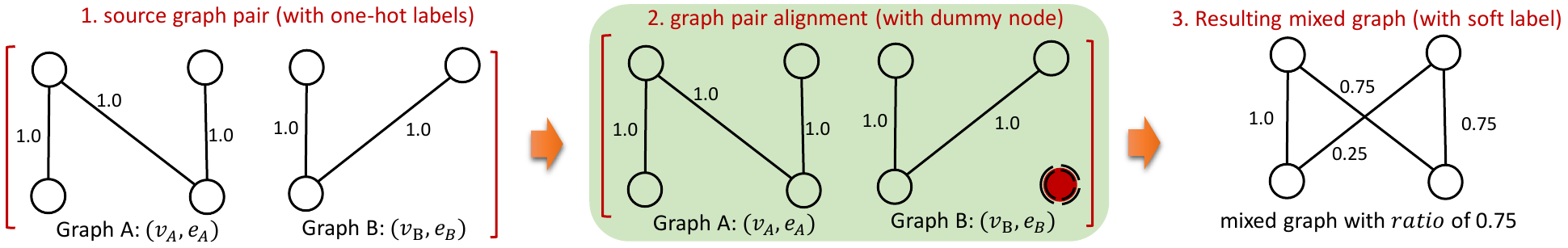}}		
	\vspace{-2mm}
	\caption{The proposed mixing schema. The left figure depicts the source graph pair, and the middle subfigure pictures the adding dump node process, and the right subfigure is the resulting mixed graph  with mixing ratio  $\lambda$ of 0.75.}	\label{fig:intrusionpair}
	\vspace{-2mm}
\end{figure*}

To answer these two questions, we propose a simple input mixing strategy for Mixup on  graph, coined \ifmixup\,%$ $ (\underline{\textbf{I}}nformation \underline{\textbf{F}}usion \underline{\textbf{Mixup}}),
for graph-level classification.  \ifmixup$ $ first samples random graph pairs 
  from the training data, and then  creates a synthetic graph  through mixing each selected sample pair, using a  mixing ratio 
   sampled from a Beta distribution. For  mixing, 
\ifmixup$ $ first adds disconnected dummy nodes  to make the two selected graphs have the same input size, so that it can align the two graphs with an arbitrary node order. Next, it 
simultaneously   performs linear interpolation between  the aligned node feature vectors and the aligned edge representations of the two graph inputs, through leveraging the straight-forward mixing strategy as  that of  the original Mixup~\cite{MixUp17} on image pairs. 
The newly generated mixed graphs, which have a much larger number of graphs with changing local neighborhood properties than the original training dataset, are then used for training to regularize the GNNs learning.

We conduct extensive experiments, 
using eight benchmarking tasks from  various domains, showing that our strategy can effectively regularize the graph classification  to improve its predictive accuracy, outperforming popular graph augmentation approaches and  GNN  methods. We also theoretically prove that such a simple mixing strategy processes an unique property under a mild assumption:  the mixed graph preserves all the information from its original graph pair, namely the mixing process is  information lossless.

\section{Related Work}

Most of the graph augmentation strategies are for node classification tasks, and heavily focus on perturbing nodes and edges in one given graph~\cite{10.5555/3294771.3294869,zhang2018bayesian,ChenLLLZS20,3412086zhou,10.1145/3394486.3403168,You2020GraphCL,WangWLCLH20,Fu2020TSExtractorLG,abs-2009-10564,abs-2104-02478,zhao2021data,LiuZXL022,zhao2022graph,arxiv.2202.08235,arxiv.2204.10390}. For example, DropEdge~\cite{rong2020dropedge} randomly removes a set of edges of a given graph. To improve the performance of temporal graphs, ~\citet{NEURIPS2021_0b0b0994} devise an efficient adaptive data augmentation, where  a few graphs are generated with different data augmentation magnitudes, and then message passing is performed between
these graphs. 
  Unlike these approaches, our proposed strategy leverages  a  pair of graphs, instead of one graph,  to augment the learning of  graph level classification. 

Despite its great success in 
augmenting data  for image  recognition and text understanding~\cite{MixUp17,guo2019,ManifoldMixUp,GuoMZ19,Guo_2020,jindal-etal-2020-augmenting,kimICML20}, Mixup has been less explored for graph learning. 
GraphMix~\cite{DBLP:journals/corr/abs-1909-11715} leverages the idea of mixing on the embedding layer,  for graph node classification in semi-supervised learning. 
MixupGraph~\cite{mixupgraph} also leverages a  simple way to avoid dealing with the arbitrary structure in the input space for mixing a graph pair, through mixing  the  graph representation resulting from passing the graph through the GNNs. Similarly, a concurrent work  G-Mixup~\cite{arxiv.2202.07179} 
first interpolates represented graph generators (i.e., graphons) of different classes, and then leverages the mixed graphons for sampling to generate synthetic graphs. 

To the best of our knowledge, 
there are only a few attempts on mixing graph inputs
~\cite{ParkSY22,arxiv.2202.10107}. 
These methods, however, rely on cumbersome processes  to first select subgraphs from the given graph pair and then design new edges to connect these subgraphs to form the mixed graph. Compared with these  works, our proposed input mixing strategy is much simpler, following the straight-forward mixing idea from that of the original Mixup~\cite{MixUp17} paper. 
Furthermore,  the mixed graphs in our method can preserve all the information from its input pairs.

\section{Preliminaries}

\textbf{Graph Classification with GNNs} In the graph classification problem we consider, the input $G$ is a graph labelled with node features, or a ``node-featured graph''. Specifically, assume that the graph has $n$ nodes, each identified with an integer ID in $[n]:=\{1, 2, \ldots, n\}$. The set of edges $E$ of the graph is a collection of unordered pairs $(i, j)$'s of node-IDs, as the current paper considers only undirected graphs (although there is no difficulty to extend the setting to directed graphs). Associated with node $i$, there is a feature vector $v(i)$ of dimension $d$. We will use $v$ to denote the collection of all feature vectors and one may simply regards $v$ as a matrix of dimension $n\times d$. Thus, each input node-featured graph $G$ is essentially specified via the pair $(v, E)$. The output $y$ is a class label in a finite set ${\cal Y}:=\{1, 2, \ldots, C\}$, which will be expressed  as a (one-hot) probability vector over the label set ${\cal Y}$. The classification problem is to find a mapping that predicts the label $y$ for a node-featured graph $G$. The training data $\mathcal{G}$ of this learning task is a collection of such $(G, y)$ pairs .

Modern GNNs use the graph structure and node features to learn a distributed vector to represent a graph. The learning  follows the ``message passing'' mechanism for neighborhood aggregation. 
It iteratively updates the embedding of a node  $h_{v}$ by aggregating representations/embeddings of its neighbors. The entire graph representation $h_{G}$ is  then obtained through a READOUT function, which  aggregates embeddings from all nodes of the graph. 
Formally,   representation $h_{i}^{k}$ of node $i$ at the $k$-th layer of a GNN is defined as: 
\begin{equation}
\label{gcnAgg}
  h_{i}^{k} = \text{AGGREGATE} (h_{i}^{k-1}, {h_{j}^{k-1}|j \in \mathcal{N}(i)}, W^{k}) , 
\end{equation}
where  $W^{k}$ denotes the trainable weights at layer $k$, $\mathcal{N}(i)$ denotes the set of all nodes adjacent to $i$, and \text{AGGREGATE} is an aggregation function implemented by  specific GNN model (popular ones include Max, Mean, Sum pooling operations), and $h_{i}^{0}$ is typically initialized as the input node feature $v(i)$ . 
The graph representation $h_{G}$ aggregates node representations $h_{v}$  using the READOUT graph pooling function:
\begin{equation}
\label{readout}
    h_{G} = \text{READOUT}({h_{i}^{k}|i \in [n]}). 
\end{equation}
This graph representation is then mapped to  label $y$ using a standard classification network (e.g., a softmax layer). 

\vspace{2mm}

\noindent \textbf{Mixup Interpolation for Images} 
Mixup  was introduced by~\cite{MixUp17}   as an interpolation-based regularizer for  image classification. It regularizes the learning of deep classification models by training with  synthetic samples, which are created by  linearly  interpolating  a pair of randomly selected training samples as well as their modeling targets. 
In detail, let $(x_A, y_A)$ and $(x_B, y_B)$ be two training instances, in which $x_A$ and $x_B$ refer to the input images  and $y_A$ and $y_B$ refer to their corresponding labels.
For a randomly chosen such training pair, 
Mixup generates a synthetic sample as follows.
\begin{equation}
\widetilde{x} = \lambda x_{A} + (1- \lambda ) x_{B},
\label{eq:mixX}
\end{equation}
%\vspace{-3mm}
\begin{equation}
\label{eq:mixY}
\widetilde{y} = \lambda y_{A} + (1- \lambda ) y_{B},
\end{equation}
where $\lambda$ is a scalar mixing ratio, sampled from a Beta($\alpha, \beta$) distribution with  hyper-parameters $\alpha$ and $\beta$.  Such synthetic instances $(\widetilde{x}, \widetilde{y})$'s are then used for training.

Motivated by the simplicity and effectiveness of Mixup in regularizing image classification models, we are naturally motivated to design a similar ``Mixup'' scheme for graph data, in particular, the node-featured graphs, as are the interest of this paper. When this is possible, we may use the synthetic instances $(\widetilde{G}, \widetilde{y})$'s to learn the model parameter  $\theta$ by minimizing the loss $\mathcal{L}$: 
\begin{equation}
    \min_{\theta} \mathbb{E}_{(G_{A}, y_{A})  \sim \mathcal{G}, (G_{B}, y_{B}) \sim \mathcal{G}, \lambda \sim Beta (\alpha, \beta)} 
    \lceil \mathcal{L} (\widetilde{G}, \widetilde{y}) \rceil.
\end{equation}

To mix $(G_A, y_A)$ and $(G_B, y_B)$, it is straight-forward to apply Equation (\ref{eq:mixY}) to obtain the mixed label $\widetilde{y}$. The key question of investigation is how to mix $G_A$ and $G_B$ to obtain $\widetilde{G}$. We will answer this question next.

\section{The Proposed Method}

We here propose a simple approach, ifMixup, for generating mixed node-featured graph $\widetilde{G}$ from a pair of training  graphs $G_A$ and $G_B$.

\subsection{Graph  Mixing Strategy}
\label{mixschema}

Given a node featured graph $G=(v, E)$, we represent $E$ as a binary matrix $e$ with $n$ rows and $n$ columns, in which $e(i, j)=1$ if $(i, j)\in E$, and $e(i, j)=0$ otherwise. Thus instead of expressing $G$ as $(v, E)$ we express it as $(v, e)$. The mixing of $G_A=(v_A, e_A)$ with $G_B=(v_B, e_B)$ to obtain $\widetilde{G}=(\widetilde{v}, \widetilde{e})$  can simply be done as follows:
\begin{equation}
\label{edgemix}
 \widetilde{e}=\lambda e_A+ (1- \lambda) e_B.   
\end{equation}
\begin{equation}
\label{graphmixingNode}
\widetilde{v} = \lambda v_{A} + (1- \lambda ) v_{B}.
\end{equation}
In order for the above mixing rule to be well defined, we need the two graphs to have the same number of nodes. For this purpose we define $n=\max(n_A, n_B)$, where $n_A$ and $n_B$ are the number of nodes in instances $A$ and $B$ respectively. If $G_A$ or $G_B$ has less than $n$ nodes, we simply introduce dummy node to the graph and make them disconnected from the existing nodes. The feature vectors for the dummy nodes are set to the all-zero vector.

This mixing process is illustrated 
in  Figure~\ref{fig:intrusionpair}, 
where the left is the source graph pair, and the middle depicts the added dummy node (i.e., the node with  dotted double circles). The right is the resulting mixed graph, where edge weights sit next to their edges. 

It is worth noting that the resulting mixed graphs, through Equations~\ref{edgemix} and~\ref{graphmixingNode}, contain  edges with weights between [0, 1]. As a result, during training, this will require the GNN networks be able to take the edge weights into account for message passing. 
Fortunately,  the two  popular GNN networks, namely GCN~\cite{kipf2017semi} and GIN~\cite{xu2018how}, can naturally cope with  weighted edges in their implementation of Equation~\ref{gcnAgg}. 

GCN handles edge weights naturally by enabling adjacency matrix to  have values between zero and one~\cite{kipf2017semi}, instead of either zero or one, representing edge weights:
\begin{equation}
\mathbf{h}^{k}_i = \sigma \left( W^k \cdot \left( \sum_{j \in \mathcal{N}(i) \cup
\{ i \}} \frac{e(i,j)}{\sqrt{\hat{d}_j \hat{d}_i}} \mathbf{h}^{k-1}_{j} \right) \right),
\end{equation}
where  $\hat{d}_i = 1 + \sum_{j \in \mathcal{N}(i)} e(i, j)$; $W^{k}$ stands for the trainable weights at layer $k$; $\sigma$ denotes the non-linearity transformation, i.e. the ReLu function. 

To enable GIN to handle soft edge weight, we replace the sum operation of the isomorphism operator in GIN with a weighted sum calculation. That is, the GIN updates  node representations as: 
\begin{equation}
\label{ginagg}
\mathbf{h}^{k}_i = \text{MLP}^{k} \left( (1 + \epsilon^{k}) \cdot
\mathbf{h}_i^{k-1} + \sum_{j \in \mathcal{N}(i)} e(i, j) \cdot \mathbf{h}_j^{k-1} \right),
\end{equation}
where  $\epsilon^{k}$ is a learnable parameter. 
The pseudo-code of  \ifmixup$ $ is described in Algorithm~\ref{alg:mixup}.
\small
\begin{algorithm}[tb]
   \caption{The mixing schema in \ifmixup}
   \label{alg:mixup}
   \textbf{Input}:  a graph pair 
   $G_A=(v_A, e_A)$ and  $G_B=(v_B, e_B)$  
   (all edges in  $e$ have weight 1)
   \textbf{Parameter}: mixing ratio $\lambda \in (0, 1)$ 
\textbf{Output}:  a mixed graph $\widetilde{G}=(\widetilde{v}, \widetilde{e})$
\begin{algorithmic}[1]
   \STATE  Compute max node number, $n=\max(n_A, n_B)$
%\For {$(u, v) \in  E$}
\IF{$ n_A < n$}
   \STATE  Add $n- n_A$ dummy nodes to $G_A$
\ELSIF{$ n_B < n$}
   \STATE  Add $n- n_B$ dummy nodes  to $G_B$
\ENDIF
 %\EndFor  
\STATE $ \widetilde{e}=\lambda e_A+ (1- \lambda) e_B   $
\STATE $\widetilde{v} = \lambda v_{A} + (1- \lambda ) v_{B}$
\STATE \textbf{return} mixed graph $\widetilde{G}=(\widetilde{v}, \widetilde{e})$
\end{algorithmic}
\end{algorithm}
\normalsize

\subsection{Information Lossless Property}
\label{sec:discu}
In this section, we show that a mixed sample generated by  \ifmixup$ $ preserves all the information from its original sample pair used for mixing, namely being information lossless. 

Such information lossless is owning to the fact that  the mixing strategy in \ifmixup$ $  guarantees that the original two node-featured graphs $G_A$ and $G_B$ recoverable from the mixed graph $\widetilde{G}$ under a mild assumption. 
In detail, to be able to recover the original graphs $G_A$ and $G_B$,  
both the {\em graph topology} and the {\em node features} of the $G_A$ and $G_B$ graphs need be recovered from the mixed instance $\widetilde{G}$. We refer to the former as {\em edge invertibility} and the latter as {\em node feature invertibility}, as will be discussed in detail next. 

\begin{lem}[\textbf{Edge Invertibility}]
 Let $\widetilde{e}$ be constructed using
 Equation \ref{edgemix} with $\lambda\neq 0.5$. Consider equation 
 \[
 s e + (1-s) e' = \widetilde{e} 
 \]
 with unknowns $s$, $e$ and $e'$,  where $s$ is a scalar and $e$ and $e'$ are binary (i.e., $\{0, 1\}$-valued) $n\times n$ matrices. There are exactly two solutions to this equation:
\[
\left\{
\begin{array}{lll}
s=\lambda,  & e= e_A,  & e'= e_B, {\rm or} \\
s=1-\lambda,  & e=e_B,  & e' = e_A\\
\end{array}
\right.
\]
\label{lem1}
 \end{lem}
 %\vspace{-5mm}
 %Section~\ref{proofedge}.
 
By this lemma, we see that if the mixing coefficient $\lambda \neq 0.5$, from the mixed edge representation $\widetilde{e}$, we can always recover $e_A$ and $e_B$ (and hence $E_A$ and $E_B$) and their corresponding weights used for mixing.  Note that if $\lambda$ is drawn from a continuous distribution over $(0, 1)$, the probability it takes value $0.5$ is zero. That is,  the connectivity of the original two graphs can be perfectly recovered from the mixed edge representation $\widetilde{e}$.
In other words, from $\widetilde{G}$, \ifmixup$ $ can perfectly recover the topologies (edge connections) of $G_A$ and $G_B$.

\begin{lem}[\textbf{Node Feature Invertibility}] Suppose that the node feature vectors for all instances in the task take values from a finite set $V\subset {\mathbb R}^d$ and that $V$ is linearly independent. Let $\widetilde{v}$ be constructed using Equation \ref{graphmixingNode}.
Let $V^*=V \cup \{{\bf 0}\}$, where ${\bf 0}$ denotes the zero vector in ${\mathbb R}^d$.
Consider equation 
\[
\widetilde{v}= s v+ (1-s)v'
\]
in which $n\times d$ matrices $v$ and $v'$ are unknowns with rows taking value in $V^*$. For any fixed $s\in (0, 1)$, there is exactly one solution of $(v, v')$ for this equation.
\label{lem2}
\end{lem}

By this lemma, we see that if the node \textit{\textbf{feature set}} $V$ is linearly independent, \ifmixup$ $ can perfectly recover the node  features of the two original graphs from the mixed graph. In other words, from $\widetilde{G}$, \ifmixup$ $ can  recover all the node features for $G_A$ and $G_B$.

In summary, these two lemmas state that, 
%Lemmas~\ref{lem1} and~\ref{lem2} state that,
given a  mixed graph $\widetilde{G}$ \ifmixup$ $ can perfectly recover the original two node-featured graphs $G_A$ and $G_B$, both their graph topologies and all the node features. %Such invertibility   indeed prevents ifMixup from manifold intrusion, as will be discussed next. 
Such invertibility makes sure that a mixed graph in \ifmixup$ $ preserves all the information from its original sample pair, being information lossless.

%Note that, the proofs for the above two lemmas are provided in the Appendix.
%\vspace{-2mm}

\subsection{Discussion}

\subsubsection{Manifold Intrusion}
The manifold intrusion degrades the performance of  Mixup-like methods~\citep{GuoMZ19,arxiv.2202.07179}. It  refers to  a form of under-fitting, resulting from conflicts between the labels of the synthetic examples and the labels of original training data~\citep{GuoMZ19}. Under the context of graph classification, the manifold intrusion  represents that the generated graphs have identical topology but 
different label than some original graphs.

\begin{theorem}[\textbf{Intrusion-Freeness}]
\label{thm:noIntrusion}
Suppose that $\lambda\neq 1/2$ and that  the condition for Lemma~\ref{lem2} is satisfied. 
% or the condition for Lemma 3 is satisfied.
Then for any mixed node-featured graph 
$\widetilde{G}=(\widetilde{v}, \widetilde{e})$ constructed using Equations \ref{edgemix} and \ref{graphmixingNode}, the two original node-feature graph $G_A$
 and $G_B$ can be uniquely recovered. 
\end{theorem}
%}

\noindent {\em Proof:} Since $\lambda\neq 0.5$,  by Lemma~\ref{lem1} we can recover $e_A$, $e_B$ and $\lambda$ from $\widetilde{e}$. Given $\lambda$ and  the condition for Lemma~\ref{lem2}, we can recover $v_A$ and $v_B$ from $\widetilde{v}$.  \hfill $\Box$%}

By this theorem, there is no {\em other pair} $(G'_A, G'_B)$ (different from $G_A$ and $G_B$) from the training set  $\mathcal{G}$ that can be mixed into $\widetilde{G}$ using any $\lambda$ due to the {\em uniqueness} in the invertibility. 
That is, it is 
impossible that 
   the mixed graph $\widetilde{G}$ to coincide  with another mixed graph $\widetilde{G'}$ resulting from any other graph pairs or with any other graphs in the training set $\mathcal{G}$ (which has one-hot label). 
Thus, manifold intrusion does not occur 
under the theorem.  

\subsubsection{Impact of Node Alignment in Mixing} 
It is worth noting that, \ifmixup$ $ mixes different graphs that are not naturally well-aligned, which seems ``counter-intuitive". Nevertheless, such  ``counter-intuition" technique has been proved to work
very well for augmenting images and texts, where Mixup has been
successfully deployed to mix two unrelated images~\cite{MixUp17,ThulasidasanCBB19} or 
two semantically different words in sentences~\cite{guo2019,Guo_2020}. 
Also, to this end, \ifmixup$ $ leverages an arbitrary node order for mixing after aligning the two input graphs by adding  dummy nodes. As a result, different node order for the same input graph pair will result in a different mixed graph.  
We  empirically evaluate the impact of such node alignment in the Ablation section in the experiments.

\subsubsection{Impact of Graph Size for Mixing} 
\ifmixup$ $ leverages adding dummy nodes to align the input graph pair, which is essentially a  padding-based strategy. In this sense, the size of the resulting mixed graph, through mixing two graphs with large difference in node numbers, can be  different from that of its source input pair.  
This is also expected to impact \ifmixup's predictive performance. We  also conduct experiments to evaluate  such effect in the Ablation section.

\subsubsection{Graph with Weighted Edges and Edge Features} 
\ifmixup$ $ assumes that the given graphs for training have binary edges, which is a widely adopted setting in learning from topological graphs.  
In the applications of graphs with real-valued edge weights,  
two main strategies can be used by \ifmixup$ $ to cope with such situation in the literature. One common strategy is to add the  edge weights to the features of the nodes~\citep{abs-2005-00687,DBLP:journals/corr/abs-2006-07739}. Another approach is to treat the  edge weights
 the same way as node features in the Aggregation function of the GNNs~\citep{DBLP:journals/corr/GilmerSRVD17,xu2018how,kipf2017semi,hu2020strategies}. It is also worth noting  that the above two approaches can also be used by \ifmixup$ $ to handle graphs that come with  edge attributes.

%\textcolor{red}{\textbf{Node Alignment and Impact}}

%\vspace{-3mm}
\section{Experiments}
\subsection{Settings}
\label{setting}
\textbf{Datasets}
We evaluate our  method with eight  graph classification tasks from the  graph benchmark datasets  collection TUDatasets~\cite{Morris+2020}:  PTC\_MR, NCI109, NCI1, and MUTAG for small molecule classification, ENZYMES and PROTEINS for protein categorization, and  IMDB-M and IMDB-B for social networks classification. 
 These  datasets have been widely used for benchmarking such as in~\cite{xu2018how} and 
can be downloaded directly using PyTorch Geometric~\cite{Fey/Lenssen/2019}'s build-in function online~\footnote{https://chrsmrrs.github.io/datasets/docs/datasets}. 
The  social networks datasets IMDB-M and IMDB-B have no node features, and we use the node degrees as  feature as  in~\cite{xu2018how}. 
%Data statistics of these datasets are shown in Table~\ref{tab:data} in the Appendix, including the number graphs, the average node number per graph,  the average edge number per graph, the number of node features, and the number of classes. 

\iffalse
\begin{table}[h]
  \centering
\scalebox{0.95}{
\begin{tabular}{l|l|c|c|c|c}\hline
Name &	graphs&	nodes &	edges& features& classes	\\ \hline

PTC\_MR&	334	&14.3&	29.4&	18&	2		\\
NCI109&	4127&	29.7&	64.3&	38&	2		\\
NCI1&	4110&	29.9&	64.6&	37&	2		\\
MUTAG	&188&	17.9&	39.6&	7&	2		\\
ENZYMES	&600	&32.6&	124.3&	3&	6		\\
PROTEINS&	1113&	39.1&	145.6&	3&	2 \\
IMDB-M&	1500	&13.0	&65.9& N/A	&	3 \\
IMDB-B&1000&19.8&96.5&N/A&2\\
\hline
\end{tabular}
}
\caption{Statistics of the graph classification  datasets. 
  }   
  \label{tab:data} 
\end{table} 
\fi

It is worth noting  that, 
the node features  of all these eight established benchmark datasets   are  {\em one-hot coded} as implemented 
by the built-in torch\_geometric.datasets functions  in the Pytorch Geometric platform. As a result, the node {\em feature sets}  of all these eight popular benchmark datasets are {\em linearly independent},  satisfying the mild assumption as stated in Lemma~\ref{lem2} for \ifmixup$ $ to be  completely information lossless. %We  further elaborate this in the Appendix. % Section~\ref{assumptionLinear}. 
%}

\begin{table*}[ht]
  \centering
   \scalebox{0.95}{
\begin{tabular}{l|c|c|c|c|c|c|c}\hline
&GCN	Baseline&		ifMixup	&	MixupGraph 	&	DropEdge 	&	DropNode	&	Attr. Masking 	&Rel. Impr. \\ \hline
PTC\_MR	&0.621$\pm$0.018&0.654$\pm$0.003&0.633$\pm$0.012&0.653$\pm$0.007&0.648$\pm$0.018&\textbf{0.666$\pm$0.009}&5.31\%\\
NCI109	&0.803$\pm$0.001&\textbf{0.820$\pm$0.005}&0.801$\pm$0.005&0.801$\pm$0.001&0.793$\pm$0.015&0.772$\pm$0.002&2.12\%\\
NCI1	&0.804$\pm$0.005&\textbf{0.819$\pm$0.004}&0.808$\pm$0.004&0.811$\pm$0.002&0.805$\pm$0.019&0.789$\pm$0.001&1.87\%\\
MUTAG	&0.850$\pm$0.011&\textbf{0.879$\pm$0.003}&0.860$\pm$0.006&0.855$\pm$0.008&0.829$\pm$0.006&0.806$\pm$0.004&3.41\%\\
ENZYMES	&0.541$\pm$0.001&\textbf{0.570$\pm$0.014}&0.551$\pm$0.016&0.566$\pm$0.006&0.532$\pm$0.006&0.506$\pm$0.021&5.36\%\\
PROTEINS&0.742$\pm$0.003&\textbf{0.753$\pm$0.008}&0.742$\pm$0.003&0.750$\pm$0.003&0.748$\pm$0.001&0.748$\pm$0.005&1.48\%\\
IMDB-M &0.515$\pm$0.002&\textbf{0.523$\pm$0.004}&0.513$\pm$0.003&0.514$\pm$0.00.&0.512$\pm$0.003&0.514$\pm$0.003&1.55\%\\
IMDB-B&0.758$\pm$0.004&\textbf{0.763$\pm$0.003}&0.759$\pm$0.002&0.762$\pm$0.004&$0.761\pm$0.005&0.756$\pm$0.003&0.66\%\\
\hline
\end{tabular}
}
\caption{Accuracy  of the testing methods with GCN (with Skip Connection) networks as baseline.  We report mean accuracy over 3 runs of 10-fold cross validation with standard deviations (denoted $\pm$). The relative improvement of ifMixup over the baseline GCN is  provided in the last  column of the table. Best results are  in \textbf{Bold}. 
  }  
  	 \vspace{-3mm}
  \label{tab:accuracy:gcn} 
\end{table*} 

\begin{table}[ht]
  \centering
   \scalebox{0.93}{
\begin{tabular}{l|c|c|c}\hline
&GAT&GATv2	&		\ifmixup \\ \hline
PTC\_MR	&0.635$\pm$0.020&0.647$\pm$0.002&\textbf{0.654$\pm$0.003}\\
NCI109	&0.769$\pm$0.005&0.768$\pm$0.007&\textbf{0.820$\pm$0.005}\\
NCI1	&0.784$\pm$0.004&0.791$\pm$0.014&\textbf{0.819$\pm$0.004}\\
MUTAG	&0.831$\pm$0.002 &0.835$\pm$0.010&\textbf{0.879$\pm$0.003} \\
ENZYMES	&0.468$\pm$0.011&0.477$\pm$0.021&\textbf{0.570$\pm$0.014}\\
PROTEINS&0.743$\pm$0.002&0.735$\pm$0.010&\textbf{0.753$\pm$0.008}\\
IMDB-M &0.514$\pm$0.004&0.511$\pm$0.001&\textbf{0.523$\pm$0.004}\\
IMDB-B&0.755$\pm$0.017 &0.753$\pm$0.002&\textbf{0.763$\pm$0.003}\\
\hline
\end{tabular}
}
  %	 \vspace{-3mm}
\caption{Comparison of GAT, GATv2, and \ifmixup$ $ with GCN  as baseline.  We report mean accuracy over 3 runs of 10-fold cross validation with standard deviations (denoted $\pm$). The relative improvement of ifMixup over the baseline GCN is  provided in the last  column. Best results are  in \textbf{Bold}. 
  }  
   \vspace{-3mm}
  \label{tab:accuracy:gcn22} 
\end{table} 

\vspace{1mm}
\noindent\textbf{Comparison Baselines} 
We compare our method with seven baselines, including  MixupGraph~\cite{mixupgraph},  DropEdge~\cite{rong2020dropedge}, DropNode~\cite{10.5555/3294771.3294869,abs-1801-10247,10.5555/3327345.3327367}, Attribute Masking and Baseline. 
MixupGraph 
applies 
Mixup on graph classification. It  leverages a  simple way to avoid dealing with the arbitrary structure for mixing a graph pair, through mixing  the entire graph representation resulting from the READOUT function of the GNNs. DropEdge and DropNode are two widely used graph perturbation strategies for   graph augmentation. DropEdge randomly removes a set of  existing edges from a given graph. DropNode randomly deletes a portion of nodes  and their connected edges.
Attribute Masking represents one of the state-of-the-art  methods for graph data augmentation~\cite{You2020GraphCL}. We here assign random  attribute values to 20\% of the graph nodes. 
Also, since \ifmixup$ $ utilizes weighted edges, we  also further compare our method with   two  state-of-the-art  attention-based graph networks:  GAT~\citep{velickovic2018graph} and GATv2~\citep{brody2022how}, which leverage learned edge weights for neighbour aggregation.  For these two methods, we also used the implementation from the Pytorch Geometric~\citep{Fey/Lenssen/2019}  platform.

For the Baseline model, we use two popular GNNs network architectures: GCN~\cite{kipf2017semi} and GIN~\cite{xu2018how}. 
GCN and GIN are  two  popular GNN  architectures and have been widely adopted for graph classification.  GCN leverages spectral-based convolutional operation to learn spectral features of graph through aggregation, benefiting from a normalized adjacency matrix, while GIN leverages  the nodes' spatial relations to aggregate neighborhood features, representing the state-of-the-art GNN network architecture. 
We use their implementations in the PyTorch Geometric platform~\footnote{https://github.com/pyg-team/pytorch\_geometric}. Note that, for the GCN, we use the GCN with  Skip Connection~\cite{7780459} as that in~\cite{li2019deepgcns}, This Skip Connection empowers the GCN to benefit from deeper  GNN networks.

\begin{figure}[h]
	\centering
	{\includegraphics[width=0.406\textwidth]{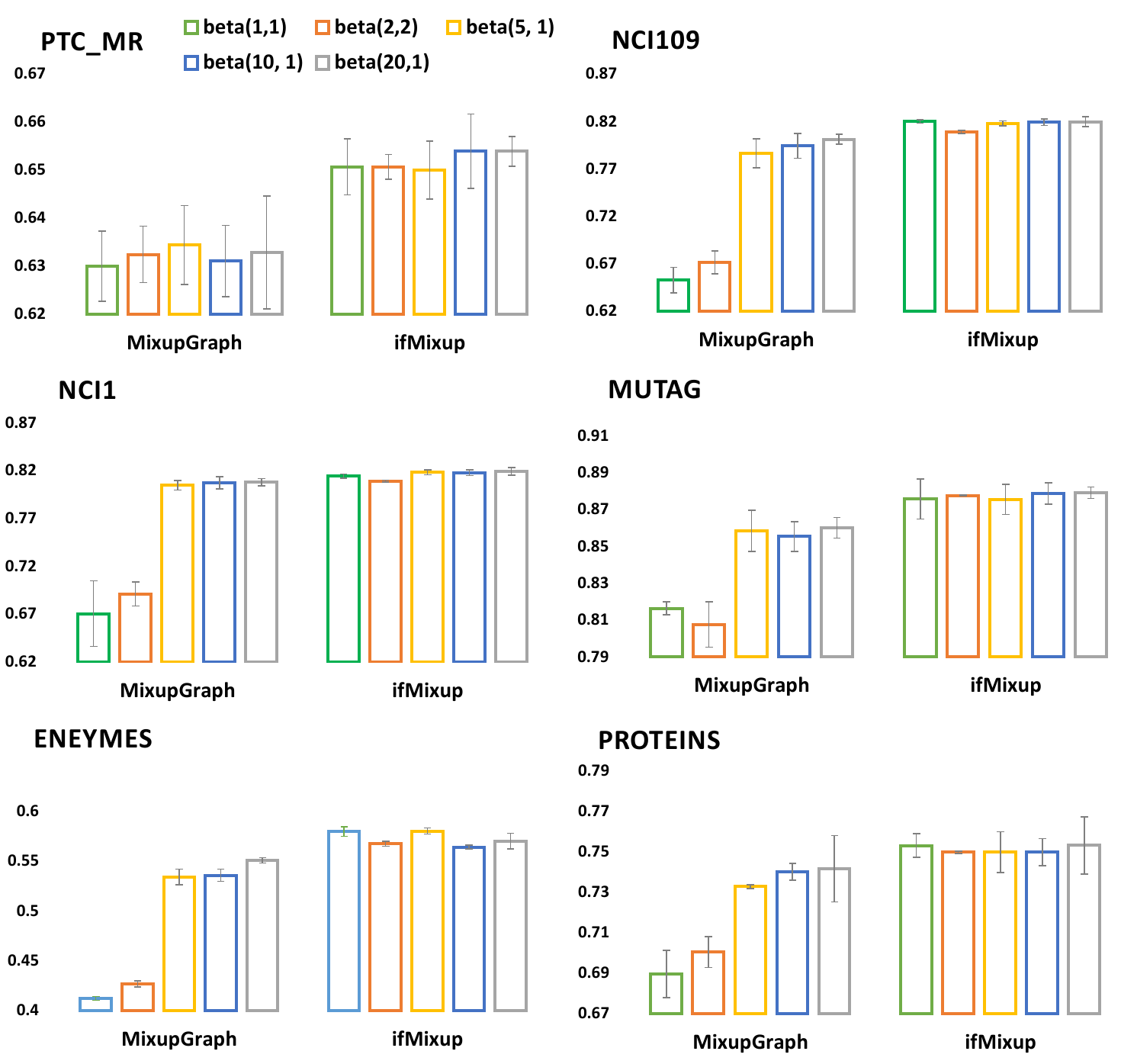}}	
		\vspace{-2mm}
	\caption{Accuracy obtained by MixupGraph and \ifmixup$ $ with mixing ratios  sampled from   Beta(1, 1), Beta(2, 2), Beta(5, 1), Beta(10, 1) and Beta(20, 1) on the six benchmark datasets.}
	\vspace{-3mm}
	\label{fig:beta}
\end{figure}

\vspace{1mm}
\textbf{Detail Settings} 
We follow the evaluation protocol and 
hyperparameters search of GIN~\cite{xu2018how} and DropEdge~\cite{rong2020dropedge}. We evaluate the models using 10-fold cross validation, and  
 report the mean and standard deviation of three  runs  on a  cluster with GPU nodes of NVidia V100 with 32 GB memory. 
 Each fold is trained with 350 epochs with AdamW optimizer~\cite{KingmaB14}, and the initial learning rate  is reduced by half every 50 epochs. 
 The hyper-parameters we search for  all models on each dataset are as follows:  (1) initial learning rate $\in$ \{0.01, 0.0005\}; 
 (2) hidden unit of size $\in$ \{64, 128\}; (3)  batch size $\in$ \{32, 128\}; (4)  dropout ratio  after the dense layer $\in$ \{0, 0.5\};  (5) DropNode and DropEdge drop ratio $\in$ \{20\%, 40\%\}; (6) number of layers in GNNs $\in$ \{3, 5, 8\}; (7)  Beta distribution for ifMixup, MixupGraph and Manifold Mixup $\in$ \{Beta(1, 1), Beta(2, 2), Beta(20, 1)\}. 
 Following GIN~\cite{xu2018how} and DropEdge~\cite{rong2020dropedge}, we  report the case giving
 the best 10-fold average cross-validation accuracy.

\subsection{Results of  GCN (with Skip Connection)  as baseline}
The accuracy obtained by the GCN (with Skip Connection) baseline, ifMixup, MixupGraph,  DropEdge,  DropNode and Attribute Masking  with GCN (with Skip Connection)
 on the eight  datasets are presented in Table~\ref{tab:accuracy:gcn} (best results  in \textbf{Bold}).

\begin{figure}[h]
	\centering
	{\includegraphics[width=0.406\textwidth]{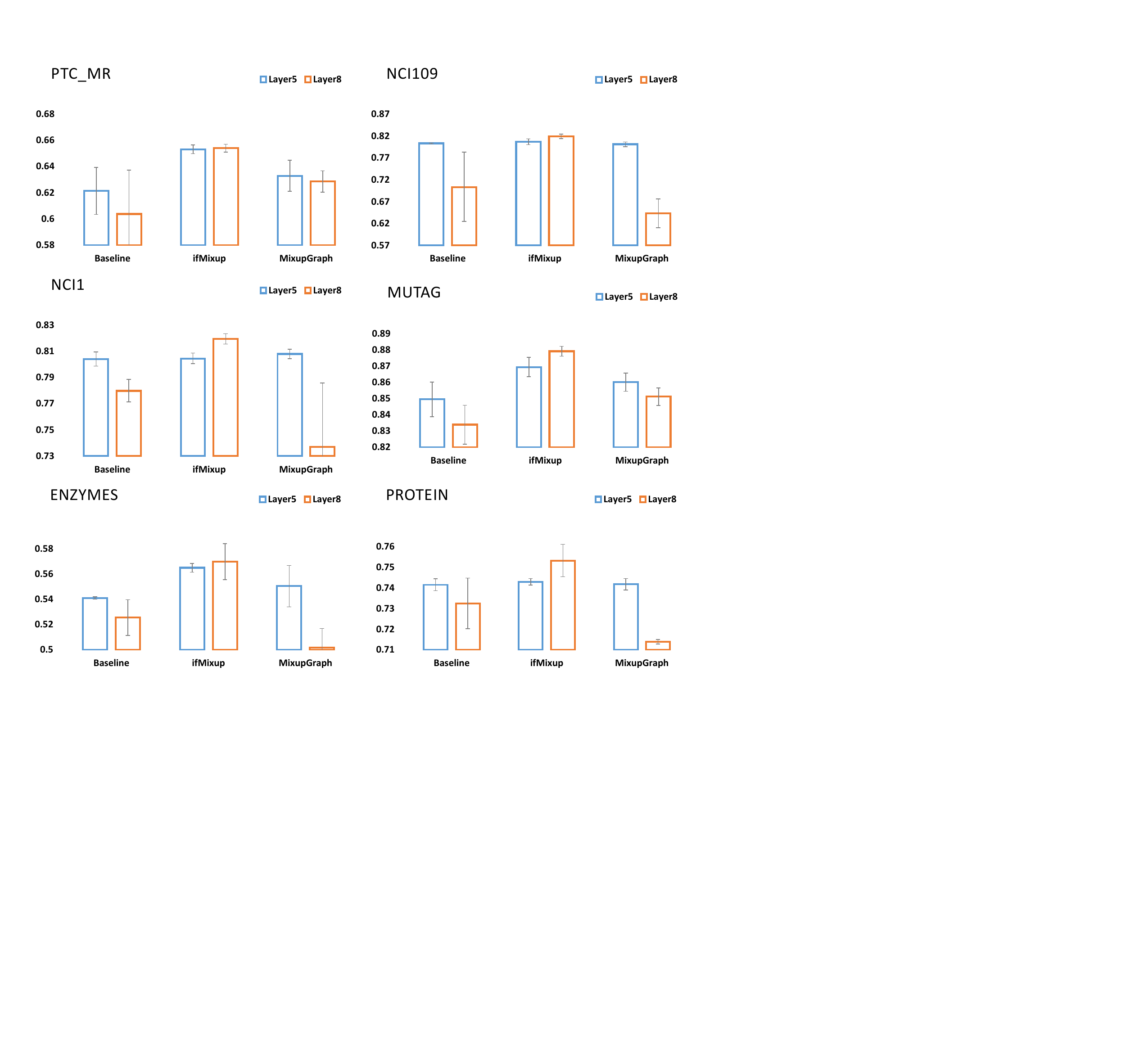}}		
	\vspace{-2mm}
	\caption{Accuracy obtained on the six benchmark datasets when varying the network depth of  GCN, \ifmixup, and MixupGraph.}
		\vspace{-4mm}
	\label{fig:layer}	
\end{figure}

Results in Table~\ref{tab:accuracy:gcn} show that ifMixup outperformed all the five comparison models against all the eight datasets, except for Attribute Masking on the PTC\_MR dataset. For example, when comparing with the GCN baseline, ifMixup obtained a relative accuracy improvement of 5.36\%, 5.31\%, and 3.41\% on the ENZYMES, PTC\_MR, and MUTAG datasets, respectively. When considering the comparison with the Mixup-like approach MixupGraph, ifMixup also obtained superior accuracy on all the eight datasets. For example, ifMixup was able to improve the accuracy over MixupGraph from 80.1\%, 80.8\%,  63.3\%, and 51.3\% to 82.0\%, 81.9\%,  65.4\%, and 52.3\% on the NCI109, NCI1,  PTC\_MR and IMDB-M datasets, respectively.

When comparing with GAT and GATv2, as shown in Table~\ref{tab:accuracy:gcn22}, \ifmixup$ $ outperformed both of the two attention-based strategies, and the improvement is with a large margin in datasets NCI09, NCI1, MUTAG, and ENZYMES.

\begin{table*}[h]
  \centering
  \scalebox{0.95}{
\begin{tabular}{l|c|c|c|c|c|c|c}\hline
&GIN	Baseline	&	ifMixup	&	MixupGraph 	&			DropEdge 	&	DropNode	&	Attr. Masking&	Rel. Impr. \\ \hline
PTC\_MR&0.644$\pm$0.007&\textbf{0.672$\pm$0.005}&0.631$\pm$0.005&0.669$\pm$0.003&0.663$\pm$0.006&0.657$\pm$	0.004
&4.35\%\\
NCI109&0.820$\pm$0.002&\textbf{0.837$\pm$0.004}&0.822$\pm$0.008&0.792$\pm$0.002&0.796$\pm$0.002&0.822$\pm$	0.002
&2.07\%\\
NCI1&0.818$\pm$0.009&\textbf{0.839$\pm$0.004}&0.822$\pm$0.001&0.791$\pm$0.005&0.785$\pm$0.003&0.825$\pm$	0.002
&2.57\%\\
MUTAG&0.886$\pm$0.011&\textbf{0.890$\pm$0.006}&0.884$\pm$0.009&0.854$\pm$0.003&0.859$\pm$0.003&0.881$\pm$	0.004
&0.45\%\\
ENZYMES&0.526$\pm$0.014&0.543$\pm$0.005&0.521$\pm$0.007&0.488$\pm$0.015&0.528$\pm$0.002&\textbf{0.544$\pm$	0.013}
&3.23\%\\
PROTEINS&0.745$\pm$0.003&\textbf{0.754$\pm$0.002}&0.744$\pm$0.005&0.749$\pm$0.002&0.751$\pm$0.005&0.748	$\pm$0.010
&1.21\%\\
IMDB-M &0.519$\pm$0.001&\textbf{0.532$\pm$0.001}&0.518$\pm$0.004&0.517$\pm$0.003&0.516$\pm$0.002&0.520$\pm$	0.004
&2.50\%\\
IMDB-B&0.762$\pm$0.004&\textbf{0.765$\pm$0.005}&0.761$\pm$0.001&0.762$\pm$0.005&0.764$\pm$0.006&0.761$\pm$	0.004
&0.39\%\\
\hline
\end{tabular}
}
\vspace{-2mm}
	 
\caption{Accuracy of the testing methods with GIN networks as baseline.  We report mean scores over 3 runs of 10-fold cross validation with standard deviations (denoted $\pm$). The relative improvement of ifMixup over the baseline GIN is   provided in the last  column of the table. Best results are  in \textbf{Bold}. 
  }  
  	\vspace{-2mm}
  \label{tab:accuracy:gin} 
\end{table*} 

\subsection{Ablation Study}
\label{ablationstu}
\subsubsection{Sensitivity of Mixing Ratio}
	 In this ablation study, we evaluate the sensitivity of the graph mixing ratio to  the two Mixup-like approaches: ifMixup and MixupGraph. We present the accuracy obtained by these two methods,   with Beta distribution as  Beta(1, 1),  Beta(2, 2), Beta(5, 1),  Beta(10, 1) and Beta(20, 1) on the %first six datasets of  Table~\ref{tab:data}.
	  six molecule datasets.  
Results are presented in Figure~\ref{fig:beta}.
Note that, the mixing ratios sampled from Beta(1, 1) follow an uniform distribution between (0, 1), and those sampled from Beta(2, 2) follow a Bell-Shaped distribution between (0, 1). Those mixing ratios have a wide range. 
On the other hand, ratios being sampled from Beta(5, 1), Beta(10, 1) and Beta(20, 1) mostly fall in the range of (0.8, 1), representing a  conservative mixing ratios.

Results in Figure~\ref{fig:beta} show that both MixupGraph  and ifMixup  obtained superior  results on the six testing datasets with Beta(20, 1). Nevertheless, MixupGraph seemed to very sensitive to the mixing ratio distribution. For example, when Beta distributions were (1, 1) and (2, 2) (first two bars in Figure~\ref{fig:beta}), MixupGraph significantly degraded its accuracy on all the six tasks (except for PTC\_MR). In contrast, ifMixup was robust to the five Beta distributions we tested.

\subsubsection{Impact of GNN Depth}
In this ablation study, we also evaluate the  accuracy obtained by  GCN, ifMixup and MixupGraph on the 
%first six datasets of Table~\ref{tab:data}, when varying the number of layers of the GCN networks.
 six molecule datasets, when varying the number of layers of the GCN networks.

The results for all the six datasets are depicted in Figure~\ref{fig:layer}. Results in this figure show that when increasing the GCN networks from 5 layers (blue bars) to 8 layers (red bars),  both GCN and MixupGraph seemed to degrade its performance on all the six datasets. For example,  for the NCI109 and NCI1 datasets, MixupGraph resulted in about 10\% of accuracy drop when increasing the number of layers in GCNs (with Skip Connection) from 5 to 8. On the contrary, ifMixup was able to increase the accuracy on all the six  datasets tested.

\begin{table}[ht]
  \centering
  \scalebox{1}{
\begin{tabular}{l|c|c}\hline
Shuffling Scenario&	NCI109 	&	  NCI1  \\ \hline
no shuffling&0.820$\pm$0.005&0.819$\pm$0.004\\
every 60 epochs&0.819$\pm$0.003&0.818$\pm$0.002\\
every 30 epochs&0.819$\pm$0.003&0.819$\pm$0.005\\
every 10 epochs&0.820$\pm$0.001&0.821$\pm$0.002\\
every 1 epoch&0.819$\pm$0.004&0.814$\pm$0.003\\
\hline
\end{tabular}
}
\vspace{-2mm}
\caption{Accuracy of  \ifmixup$ $ with various node shuffling   scenarios (with increasing shuffling frequency) for mixing.  
  }  
 % \vspace{-2mm}
  \label{tab:accuracy:shuffling} 
\end{table}

\subsubsection{Impact of Node Order}
\label{nodeorder}
In this section, we evaluate the impact of node order permutation for \ifmixup. 
We randomly shuffle the node order of one of the graphs in the graph pair before being mixing by \ifmixup.
We here vary the shuffling  frequency: shuffling  for every $k$ epochs. Here we test $k$ with 60, 30, 10, and 1 (frequency from low to high). 
We conduct experiments using 8-layer GCN on the NCI109 and NCI1 datasets, and present the results in Table~\ref{tab:accuracy:shuffling}.
Results in Table~\ref{tab:accuracy:shuffling} show that shuffling the graph for each epoch (\textit{high frequency}) may hurt the performance (e.g., for the NCI1 dataset), but \ifmixup$ $ seems to be insensitive to \textit{less frequent} shuffling scenarios, such as shuffling every 60, 30, and 10 epochs.

We suspect that the above accuracy degradation of ifMixup when shuffling graph nodes for each epoch was due to the
limit of the modeling power of the ifMixup with 8-layer GCN. % (with Skip Connection).
To verify this hypothesis, we experiment with deeper 
GCN networks for ifMixup by doubling the network depth, namely using 16 layers. The results are presented  in Table~\ref{tab:accuracy:shufflingdeeper}.

From Table~\ref{tab:accuracy:shufflingdeeper}, we can see that when increasing ifMixup's modeling capability by using deeper GCN networks, namely 
16 layers, shuffling for each epoch seemed not to degrade ifMixup's accuracy, but was able to improve  the predictive accuracy of ifMixup.  For example, ifMixup that shuffles the graph nodes for each epoch increased the accuracy from 81.4\% to 82.2\% when its depth was increased from 8 layers to 16 layers on the NCI1 dataset. As shown in Table~\ref{tab:accuracy:shufflingdeeper}, ifMixup with shuffling achieved  the best accuracy for both NCI109 and NCI1 when using 16-layer networks while shuffling the graph nodes for each epoch.  

These results suggest that, when the modeling capability is limited, frequent shuffling in ifMixup may hurt the model's predictive accuracy. This is because shuffling the graph nodes before mixing graph pairs can significantly increase the input variety, compared to that of without  shuffling. 
When we have a model with strong modeling capability, shuffling can help improve the model's predictive accuracy because the shuffling can significantly increase both the training sample size and the input variety.

\begin{table}[ht]
  \centering
  \scalebox{0.8}{
\begin{tabular}{l|l|c|c}\hline
GCN layers&Shuffling Scenario&	NCI109 	&	  NCI1  \\ \hline \hline
\multirow{ 2}{*}{8 layers}&no shuffling&\textbf{0.820$\pm$0.005}&\textbf{0.819$\pm$0.004}\\
&shuffling every epoch&0.819$\pm$0.004&0.814$\pm$0.003\\ \hline

\multirow{ 2}{*}{16 layers}&no shuffling&0.820$\pm$0.003&0.820$\pm$0.001\\
&shuffling every epoch&\textbf{0.821$\pm$0.003}&\textbf{0.822$\pm$0.004}\\
\hline
\end{tabular}
}
\vspace{-2mm}
\caption{Accuracy of  ifMixup  with and without node shuffling   vs. network depth. Best results are in \textbf{bold}.}  
 
 \vspace{-3mm}
  \label{tab:accuracy:shufflingdeeper} 
\end{table}

\begin{figure}[ht]
\vspace{-2mm}
	\centering
	{\includegraphics[width=0.48\textwidth]
	{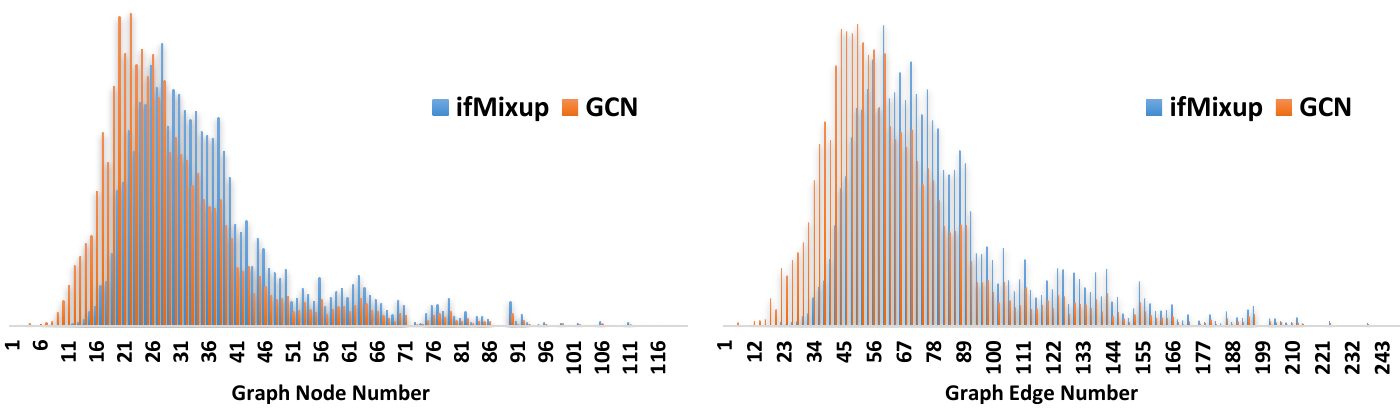}
	}	
	 \vspace{-4mm}
	\caption{The average  numbers of graph nodes (left) and edges (right)  for training  GCN and \ifmixup.}
	\label{fig:graphsize}
\end{figure}

\vspace{-3mm}
\subsubsection{Impact of Graph Size}
In  this section, we evaluate the graph size  change in training due to mixing up graph pairs by \ifmixup$ $ and its impact. 
In Figure~\ref{fig:graphsize}, we show the histograms of the number of graph nodes and edges for both the original graphs and mixed graphs on NCI109. 
From Figure~\ref{fig:graphsize}, one can see that \ifmixup$ $ \textit{slightly} shifted the node and edge distributions to the right, but with very similar distributions~\footnote{This also suggests that \ifmixup\ adds little additional computation cost to the original GNN models.}.

To further evaluate the impact of such change, we implement a variant of \ifmixup, where the pair of graphs being mixed have the same number of nodes.  We conduct experiments using 8-layer GCN on the NCI109 and NCI1 datasets, and present the results in Table~\ref{tab:accuracy:graphsize}.
From Table~\ref{tab:accuracy:graphsize}, we can see that mixing random pair outperformed the three other pairing scenarios: mixing graphs with the  same size with and without shuffling, and mixing with a shuffled version of the same graph. Interestingly, as shown in the second last two rows of the table, node shuffling seems to improve the performance when mixing random graphs with the same number of nodes.

\begin{table}[ht]
\vspace{-2mm}
  \centering
  \scalebox{1}{
\begin{tabular}{l|c|c}\hline
Mixing Pair&	NCI109 	&	  NCI1  \\ \hline\hline
random pair&0.820$\pm$0.005&0.819$\pm$0.004\\\hline
same size/no shuffling&0.787$\pm$0.003&0.795$\pm$0.002\\
same size/shuffling& 0.816$\pm$0.002&0.812$\pm$0.004\\
self-paired/shuffling&0.806 $\pm$0.003&0.807$\pm$0.003\\
\hline
\end{tabular}
}
\vspace{-2mm}
\caption{Accuracy of  \ifmixup$ $ with different pairing scenarios in terms of the number of nodes in a graph pair. Shuffling here refers to graph node order permutation.}  
 \vspace{-2mm}
  \label{tab:accuracy:graphsize} 
\end{table} 

\vspace{-2mm}
\subsection{Results of using GIN as baseline}

We also evaluate our method using the GIN~\cite{xu2018how}  network architecture. 
The accuracy obtained by the GIN baseline, ifMixup, MixupGraph, DropEdge,  DropNode, and Attribute Masking   using GIN 
as baseline on the eight test datasets are presented in Table~\ref{tab:accuracy:gin} (best results are  in \textbf{Bold}).

Table~\ref{tab:accuracy:gin} shows that, similar to the GCN case, the ifMixup with GIN as baseline outperformed all the five  comparison models against all the eight datasets, except for Attribute Masking on the ENZYMES dataset. For example, when comparing with GIN, ifMixup obtained a relative accuracy improvement of 4.35\%, 3.23\%, and 2.57\% on the  PTC\_MR, ENZYMES, and NCI1 datasets, respectively. When  comparing with the Mixup-like approach MixupGraph, ifMixup also obtained higher accuracy on all the eight datasets. For example, ifMixup was able to improve the accuracy over MixupGraph from 82.2\%, 82.2\%, 63.1\%, and 51.8\% to 83.7\%, 83.9\%,  67.2\%, and 53.2\% on the NCI109, NCI1, PTC\_MR, and IMDB-M datasets, respectively.

\vspace{-1mm}
\section{Conclusion and Future Work}

We proposed a very simple, straight-forward, input mixing strategy for Mixup on graph classification. Our method directly mixes up a pair of graph inputs, which is easy to be implemented. %and computational cheap. 
We also proved that such a simple mixing strategy processes an unique property under a mild assumption:  the mixed graph can preserve all the information from its original graph pair used for mixing. 
We  showed, using eight benchmark graph classification tasks from  different domains, that our  method obtained superior  %predictive
accuracy to popular graph augmentation approaches and  graph
classification methods. 

It is worth noting that,  
\ifmixup\ focuses on graph level classification. It becomes more challenging to guarantee  information losses for  mixed graphs on the node  level. For example, it has been well observed that GNNs  suffer from   over-smoothing~\cite{LiHW18}, where  the representations of all nodes in a graph may converge to a subspace that makes their representations  unrelated to the \textit{input}~\cite{LiHW18,abs-1901-00596}. 
That is, for node  classification, the learning dynamics of the GNNs  play a vital role on the forming of the node embeddings. 
It is   our interest to extend \ifmixup\ for graph node classification in the future.

\section{Acknowledgements}
This project was supported in part by  a National Research Council of Canada (NRC) Collaborative R\&D grant (AI4D-CORE-07).

\bibliography{graphMixup}
%\bibliographystyle{aaai23}
%\nobibliography{aaai23}

%\iffalse
\newpage
\appendix
\onecolumn

\section{Appendix}

\subsection{Proof of Lemma~\ref{lem1}(\textbf{Edge Invertibility})} 
\label{proofedge}
\noindent {\em Proof:}

First note that the values in matrix $\widetilde{e}$ can only take values in 
$\{0, \lambda, 1-\lambda, 1\}$.

The set $[n]\times [n]$ of all node pairs can be partitioned into four sets:
\begin{eqnarray*}
{\cal M}_{00} &:= &
\{(i, j) \in [n]\times [n]: e_A(i, j) = 0,  e_B(i, j) = 0\}  \\
{\cal M}_{01} &:= &
\{(i, j) \in [n]\times [n]: e_A(i, j) = 0,  e_B(i, j) =1 \}  \\
{\cal M}_{10} &:= &
\{(i, j) \in [n]\times [n]: e_A(i, j) = 1,  e_B(i, j) = 0\}  \\
{\cal M}_{11} &:= &
\{(i, j) \in [n]\times [n]: e_A(i, j) = 1,  e_B(i, j) = 1\}  
\end{eqnarray*}
It is clear that
\[
\widetilde{e}(i, j) = 
\left\{
\begin{array}{cc}
     0, & {\rm if} ~(i, j) \in {\cal M}_{00}  \\
     1-\lambda, & {\rm if} ~(i, j) \in {\cal M}_{01}  \\
     \lambda, & {\rm if} ~(i, j) \in {\cal M}_{10}  \\     
     1, & {\rm if} ~(i, j) \in {\cal M}_{11}  \\
\end{array}
\right.
\]
Let $e, e'$ and $s$ be the solution of the equation in the lemma.
On ${\cal M}_{00} \cup {\cal M}_{11}$, we must have $e=e'=\widetilde{e}$. We only need to determine $e$ and $e'$ on ${\cal M}_{01}$ and ${\cal M}_{10}$. When $\lambda \neq 0.5$, we must have either
\[
s=\lambda, 
~e(i, j) = \left\{ 
\begin{array}{cc}
1,     & (i, j) \in {\cal M}_{10}  \\
0,     & (i, j) \in {\cal M}_{01}
\end{array}
\right.
~{\rm and}~
e'(i, j) = \left\{ 
\begin{array}{cc}
0,     & (i, j) \in {\cal M}_{10}  \\
1,     & (i, j) \in {\cal M}_{01}
\end{array}
\right.
\]
or 
\[
s=1-\lambda, 
~e(i, j) = \left\{ 
\begin{array}{cc}
0,     & (i, j) \in {\cal M}_{10}  \\
1,     & (i, j) \in {\cal M}_{01}
\end{array}
\right.
~{\rm and}~
e'(i, j) = \left\{ 
\begin{array}{cc}
1,     & (i, j) \in {\cal M}_{10}  \\
0,     & (i, j) \in {\cal M}_{01}
\end{array}
\right.
\]
Comparing such solutions with $e_A$ and $e_B$, we prove the lemma. \hfill $\Box$

\subsection{Proof of Lemma~\ref{lem2} (\textbf{Node Feature Invertibility})}
\label{proofnode1}
\noindent{\em Proof:} We will prove the lemma by showing that for any $i\in [n]$, based on $v(i)$, we can uniquely recover $v(i)$ and $v'(i)$.

Case 1: $\widetilde{v}(i)={\bf 0}$. It is obvious $v(i)=v'(i)={\bf 0}$.

Case 2: $\widetilde{v}(i) \notin V$ but $\widetilde{v}=c u$ for some $u\in V$ and some scalar $c$. In this case, $c$ must be either $s$ or $1-s$. If $c=s$, then $v(i)=u, v'(i)={\bf 0}$. If $c=1-s$, then $v(i)={\bf 0}, v'(i)=u$.

Case 3: $\widetilde{v}(i) \notin V$ and 
$\widetilde{v}\neq c u$ for any $u\in V$ and any scalar $c\neq 0$.  For any two $u, u' \in V$, let ${\rm SPAN}(u, u')$ denote the vector space spanned $u$ and $u'$. Since $V$ is a linearly independent set, it is clear every choice of $(u, u')$ gives a different space ${\rm SPAN}(u, u')$, and $\widetilde{v}(i)$ must live in one and only one such space.  After identifying this space, we can identify $(u, u')$. With the knowledge of $s$, we can precisely correspond $u$ and $u'$ with $v(i)$ and $v'(i)$ since either $u=v(i)$ and $u'=v'(i)$  are true, or 
$u=v'(i)$ and $u=v(i)$ are true, but both can not be true at the same time.

Thus we have enumerated all possible cases, and in each case, there is a unique solution to the equation of interest. \hfill $\Box$

\subsection{Statistics of the Eight Benchmark Datasets Used in the Paper}
Table~\ref{tab:data} details the statistics of the 8 benchmark datasets used in the paper, including the number graphs, the average node number per graph,  the average edge number per graph, the number of node features, and the number of classes. 
\begin{table}[h]
  \centering
\scalebox{1}{
\begin{tabular}{l|l|c|c|c|c}\hline
Name &	graphs&	nodes &	edges& features& classes	\\ \hline

PTC\_MR&	334	&14.3&	29.4&	18&	2		\\
NCI109&	4127&	29.7&	64.3&	38&	2		\\
NCI1&	4110&	29.9&	64.6&	37&	2		\\
MUTAG	&188&	17.9&	39.6&	7&	2		\\
ENZYMES	&600	&32.6&	124.3&	3&	6		\\
PROTEINS&	1113&	39.1&	145.6&	3&	2 \\
IMDB-M&	1500	&13.0	&65.9& N/A	&	3 \\
IMDB-B&1000&19.8&96.5&N/A&2\\
\hline
\end{tabular}
}
\caption{Statistics of the graph classification benchmark datasets. 
  }   
  \label{tab:data} 
\end{table} 
%\fi

\end{document}